\documentclass[journal,12pt,onecolumn,draftclsnofoot,]{IEEEtran}

\usepackage[cmex10]{amsmath}
\usepackage[utf8]{inputenc}
\usepackage{multirow}
\usepackage{soul}
\usepackage{algorithm}
\usepackage{algpseudocode}
\usepackage{graphicx}
\usepackage{dsfont}
\usepackage{xcolor}
\usepackage{cite}

\newcommand{\probP}{\text{I\kern-0.15em P}}

\usepackage[utf8]{inputenc}
\usepackage{mathtools}
\usepackage[mathscr]{euscript}
\usepackage{amsmath}
\usepackage{amssymb}
\usepackage{amsfonts}
\usepackage{amssymb}
\usepackage{amsmath}
\usepackage{verbatim}
\usepackage[T1]{fontenc}
\usepackage[utf8]{inputenc}
\usepackage{latexsym}
\usepackage{enumerate}
\usepackage{indentfirst}
\usepackage{calligra}
\usepackage{ulem}
\usepackage{graphicx}
\usepackage{array}
\usepackage{float}
\usepackage{bbm}
\usepackage{color, colortbl}

\definecolor{LightCyan}{rgb}{0.88,0.75,1}	
\definecolor{Gray}{gray}{0.9}
\usepackage{geometry}

\usepackage{mleftright}
\usepackage{colortbl}
\makeatletter
\let\old@ps@headings\ps@headings
\let\old@ps@IEEEtitlepagestyle\ps@IEEEtitlepagestyle
\def\psccfooter#1{%
 \def\ps@headings{%
 \old@ps@headings%
 \def\@oddfoot{\strut\hfill#1\hfill\strut}%
 \def\@evenfoot{\strut\hfill#1\hfill\strut}%
 }%
 \def\ps@IEEEtitlepagestyle{%
 \old@ps@IEEEtitlepagestyle%
 \def\@oddfoot{\strut\hfill#1\hfill\strut}%
 \def\@evenfoot{\strut\hfill#1\hfill\strut}%
 }%
 \ps@headings%
}
\makeatother

\usepackage{soul}
\usepackage{multirow}
\usepackage{soul,color}
\usepackage{graphicx}
\usepackage{dsfont}
\usepackage{subcaption}

\usepackage[utf8]{inputenc}
\usepackage{mathtools}
\usepackage[mathscr]{euscript}
\usepackage{amsmath}
\usepackage{amssymb}
\usepackage{amsfonts}
\usepackage{verbatim}
\usepackage[T1]{fontenc}
\usepackage[utf8]{inputenc}

\usepackage{latexsym}
\usepackage{enumerate}
\usepackage{indentfirst}
\usepackage{calligra}
\usepackage{ulem}
\usepackage{multirow}

\usepackage{array}
\usepackage{float}
\usepackage{bbm}

\usepackage{eurosym}

\usepackage{hyperref}

\usepackage{caption}
\usepackage{tikz}
\usepackage{pgfplots}

\pgfplotsset{compat=1.8}
\usepgfplotslibrary{statistics}
\pgfmathdeclarefunction{fpumod}{2}{%
 \pgfmathfloatdivide{#1}{#2}%
 \pgfmathfloatint{\pgfmathresult}%
 \pgfmathfloatmultiply{\pgfmathresult}{#2}%
 \pgfmathfloatsubtract{#1}{\pgfmathresult}%
 \pgfmathfloatifapproxequalrel{\pgfmathresult}{#2}{\def\pgfmathresult{5}}{}%
 }

\usepackage{tabularx}
\usepackage{flushend}
\usepackage{textcomp}
\usepackage{xcolor}
\usepackage{import}

\usepackage{csvsimple}
\usepackage{tabularx}
\usepackage{adjustbox}

\usetikzlibrary{trees,decorations,shadows}
\tikzset{level 1/.style={sibling angle=45,level distance=4mm}}
\usetikzlibrary{arrows.meta}
\usepackage{forest}
\usetikzlibrary{external}

\let\oldtikzexternalgetnextfilename\tikzexternalgetnextfilename \renewcommand{\tikzexternalgetnextfilename}[1]{\oldtikzexternalgetnextfilename{#1}\expandafter\tikzsetnextfilename\expandafter{#1}}

\usepackage{pgfplotstable}
\usepackage[outline]{contour}
\contourlength{1.2pt}

\pgfplotsset{compat=1.13} 

\usetikzlibrary{spy}
\usetikzlibrary{calc}
\usetikzlibrary{fadings}
\usetikzlibrary{patterns}
\usetikzlibrary{shadows}
\usetikzlibrary{mindmap}
\usetikzlibrary{backgrounds}
\usetikzlibrary{shapes.symbols}
\usetikzlibrary{shapes.multipart}
\usetikzlibrary{shapes.geometric}
\usetikzlibrary{automata,positioning}
\usetikzlibrary{decorations.fractals} 
\usetikzlibrary{decorations.markings}
\usetikzlibrary{decorations.pathreplacing}
\usetikzlibrary{decorations.pathmorphing}
\usepackage{svg}
 
\usepackage{bm}

\usepackage{nomencl}
\makenomenclature
\tikzset{edge from parent/.style={segment angle=10,draw}}

\tikzset{
 my rounded corners/.append style={rounded corners=2pt},
}

\def\BibTeX{{\rm B\kern-.05em{\sc i\kern-.025em b}\kern-.08em
 T\kern-.1667em\lower.7ex\hbox{E}\kern-.125emX}}

\renewcommand{\nomgroup}[1]{%
 \ifthenelse{\equal{#1}{O}}{\item[\textit{Operators}]}{%
 \ifthenelse{\equal{#1}{I}}{\item[\textit{Indices}]}{%
 \ifthenelse{\equal{#1}{A}}{\item[\textit{Acronyms}]}{%
 `\ifthenelse{\equal{#1}{V}}{\item[\textit{Variables and parameters}]}{}}}}}
\usepackage{scalerel}
\usetikzlibrary{svg.path}
\definecolor{orcidlogocol}{HTML}{A6CE39}
\tikzset{
 orcidlogo/.pic={
 \fill[orcidlogocol] svg{M256,128c0,70.7-57.3,128-128,128C57.3,256,0,198.7,0,128C0,57.3,57.3,0,128,0C198.7,0,256,57.3,256,128z};
 \fill[white] svg{M86.3,186.2H70.9V79.1h15.4v48.4V186.2z}
 svg{M108.9,79.1h41.6c39.6,0,57,28.3,57,53.6c0,27.5-21.5,53.6-56.8,53.6h-41.8V79.1z M124.3,172.4h24.5c34.9,0,42.9-26.5,42.9-39.7c0-21.5-13.7-39.7-43.7-39.7h-23.7V172.4z}
 svg{M88.7,56.8c0,5.5-4.5,10.1-10.1,10.1c-5.6,0-10.1-4.6-10.1-10.1c0-5.6,4.5-10.1,10.1-10.1C84.2,46.7,88.7,51.3,88.7,56.8z};
 }
}

\newcommand\orcidicon[1]{\href{https://orcid.org/#1}{\mbox{\scalerel*{ \begin{tikzpicture}[yscale=-1,transform shape]
 \pic{orcidlogo};
 \end{tikzpicture}
 }{|}}}}

\begin{document}
%
\title{\huge{Fairness for distribution network operations and planning}}

\author{Pedro~F.~C.~de~Carvalho$^{1}$~\orcidicon{0009-0000-2490-7027}, ~Zijie~Liu$^{1,2}$,~Md~Umar~Hashmi$^{1,2}$,~\IEEEmembership{Senior~Member,~IEEE}~\orcidicon{0000-0002-0193-6703}, and~Dirk~Van~Hertem$^{1,2}$,~\IEEEmembership{Fellow,~IEEE}~\orcidicon{0000-0001-5461-8891}
\thanks{Corresponding author email: mdumar.hashmi@kuleuven.be}
\thanks{$^{1}$P.F.C.C, Z.L., M.U.H and D.V.H. are with KU Leuven, division Electa in Leuven }
\thanks{$^{2}$Z.L., M.U.H and D.V.H. are with \& EnergyVille in Genk, Belgium}
} 

 

 
 



\maketitle

\begin{abstract}
The incorporation of fairness into the distribution network (DN) planning and operation has become a key goal of recent studies. The cost of implementing fairness, denominated the price of fairness (PoF), covers the efficiency that is renounced for attaining social cohesion through fair outcomes. Locational disparity makes fairness schemes emerge to level the consumers playing field. However, fairness encompasses a range of notions. From egalitarian to merit-based criteria, various metrics are implemented as a tool for measuring equitable utility distribution. These have different mathematical complexities, from linear to non-linear programming cases, which affect their overall applicability. Hence, this study compiles the overarching fairness notions and metrics, reviewing how these affect stakeholders and the inherent mathematical optimisation in resource allocation problems. The aim is to support consistent and transparent planning and decision-making within DN operations. 
\end{abstract}

\begin{IEEEkeywords}
Fairness, distribution network, operation, planning, hosting capacity.
\end{IEEEkeywords}

{\textbf{Disclaimer}: This paper is a preprint of a paper submitted to and presented at the CIRED 2026 Brussels workshop on Implementing Successful Innovation in Distribution Networks, Belgium. 
The final version of the paper will be available at CIRED's website.}


\pagebreak

\tableofcontents

\pagebreak

\section{Introduction}
Fairness in distribution network (DN) planning and operations is gaining prominence, yet remains inconsistently defined. Current studies often introduce bespoke indicators and rules, limiting comparability and transparency. Moreover, the trade-off between fairness and efficiency is seldom made explicit. In fact, incorporating notions of fairness in low-voltage (LV) DNs followed the considerable growth in Distributed Energy Resources (DERs) during the last couple of decades. With Photovoltaic (PV) generation swarming both domestic and industrial infrastructure, the grid, originally designed to carry electricity from the producer to the consumer, now faces recurrent bidirectional flows. In order to avoid technical violations, congestion is managed, voltage is controlled, and the Hosting Capacity (HC), the maximum capacity of Distributed Generation (DG) that can be installed \cite{hashmi2025hosting}, is assessed. 

Fairness, more specifically unfairness, becomes an issue as the layout of the grid dictates the outcome of these prevention mechanisms. Nodes farther away from the substation suffer from locational disparity effects, receiving less HC allocation and being overall more prone to curtailments. Thus, due to power network physics, the traditional efficiency-centered resource allocation is insufficient in providing a just solution. Hence, fairness must be integrated into the optimisation problem-solving, ensuring existing disparities are not aggravated by feeder topologies.

\subsection{Literature Review and Gaps}

Among the reviewed fairness applications in DN optimisation, most studies fall under a small group of umbrella terms: Congestion Management (CM), Hosting Capacity (HC), Local Energy Systems (LESs) and Voltage Regulation (VR). 

Regarding CM, the framework presented by Liu et al. provides three different schemes, each with its own metric as a fairness proxy. The first focuses on the harvesting quantity, the second on the exports and the third on the net benefit of a given household in order to fairly implement PV curtailment  \cite{liu2020}. Abdolahinia et al. focuses on using Transactive Control Signals (TCSs) in a peer-to-peer (P2P) bi-level framework as a way to influence market participant behavior, tackling recurring DN limitations \cite{abdolahinia2025}. Hernández et al., on the other hand, propose Virtual Power Lines (VPLs) as a tool for managing congestion fairly \cite{hernandez2021}. Tackling the complications caused by rapid fluctuations in renewable energy sources, Zhan et al. propose a fairness-incorporated online feedback optimization approach \cite{zhan2024}. Liang et al. recently proposed an innovative aggregation mechanism which, as in \cite{abdolahinia2025}, implements a bi-level model as a way to ensure fairness is guaranteed throughout the process \cite{liang2025}.

While modern HC analysis also deals with load, it is traditionally focused on the maximum grid integration of DG. Rubbers et al. shed light not only on the impact topology has, comparing four different feeders, but also on different fairness notions \cite{rubbers2025}. Another example that also contrasts bounded efficiency against a fair HC allocation is Mahmoodi and Blackhall's HC Envelope study \cite{mahmoodi2021}. In fact, this study adopts one of the proportional performance metrics proposed in \cite{liu2020}, namely the PV Harvesting index (PHI). Eldridge and He propose an auction mechanism designed to coordinate DN planning by balancing consumer demand with the utility's need for infrastructure upgrades \cite{eldridge2023}. Recent studies enhance existing fairness-aware HC calculations through the use of matrices, either through convex inner approximation approaches \cite{aydin2025a}, or through sequential linear optimization methods \cite{lliuyacc-blas2025}. 

LESs include Energy Communities (ECs), Collective Self-Consumption (CSC) and P2P schemes often involving game-theory concepts. Thus, Abdolahinia et al. also contribute to these efforts by alleviating Transactive Energy Markets (TEMs) \cite{abdolahinia2025}. Soares et al. lay the groundwork on this topic by providing a comprehensive fairness review in LES studies \cite{soares2024}. For instance, two of the covered papers are \cite{casalicchio2022}, which uses an ad-hoc fairness metric, and \cite{long2019}, which uses the Shapley Value (SV) as the ideal fair distribution to achieve a representative fairness index. Subsequently, Gasca et al. assess both centralized and decentralized frameworks, highlighting the higher systematic economic yield for centralized ones \cite{gasca2025}. Furthermore, Couraud et al. provide not only a thorough description of the CSC process in France, but also a review of the main fairness types relevant to LES \cite{couraud2025}. Transversally, these studies are particularly centered around meritocratic fairness notions \cite{soares2024}.
Finally, \cite{liu2020} and \cite{hernandez2021}  also adopt PV curtailment in their frameworks as a VR tool. As some authors propose a distributed framework for fair DER coordination \cite{gebbran2021, poudel2023}, Geerdroodbari et al. focus on a decentralised control strategy approach \cite{gerdroodbari2021}. Sun et al. propose a three-stage incentive-based VR strategy, which implements a threshold-based mode selection \cite{sun2022}. Hashmi et al. evaluate locational disparity effects due to inverter injection rules by proposing a Loss of Consumer Gain (LCG) metric, serving as the fairness proxy in their study \cite{hashmi2023}. At last, Gupta and Molzahn not only aim to improve fairness in PV curtailment schemes by proposing feedback-based daily topology reconfigurations \cite{gupta2025}, but also by decomposing existing VR schemes into two groups: one that maximizes an extra fairness objective and another that employs feedback-based weights to promote fairness within the traditional curtailment minimization objective for each PV plant \cite{gupta2025a}.

Focusing on general optimization (GO), Wei and Bandi examine how to integrate fairness into network flow problems by addressing the price disparities that often emerge from purely cost-focused models \cite{wei2015}. Chen and Hooker provide one of the most comprehensive guides to formulating fairness in optimization models, covering a plethora of fairness notions, metrics and their respective Social Welfare Functions (SWFs) \cite{chen2023}. Building on that study's foundations, in \cite{sundar2025}, Sundar et al. provide a fairness representation for decision-making problems, proposing a novel metric as a solution to some critiques of the metrics presented in \cite{chen2023}. 

Hence, the gaps in existing literature consist of a lack of common fairness definitions across studies and stakeholders, without a systematic rationale for metric selection, leading to a plethora of different metrics being implemented. Additionally, there is a limited quantification of the Price of Fairness (PoF) in proposed fair frameworks, leading to a lack of transparency and a clear understanding of the true benefits of implementing fairness in DNs.

\subsection{Contributions and Paper Structure}

In this paper, we provide an overview of how fairness is integrated into the optimization models in DNs, understanding what the metrics used to represent each fairness notion are. Additionally, we also review what stakeholders are represented in existing literature, and how their needs are addressed by current fairness-aware schemes. Simultaneously, the impact of current frameworks on the PoF and locational disparity is also covered, before closing with a glance at the mathematical modeling impact of different fairness metrics and strategies. 

This paper is organized into four main sections. Section \ref{sec:stakeholders} presents which and how the different DN participants, as well as the prevalent fairness notions are represented in those works. The PoF and locational disparity effects are discussed within Section \ref{sec:pof}. Section \ref{sec:metrics} covers the essential fairness metrics as employed in DN literature, presenting their formulations, range and optimal fair value. 
Finally, Section \ref{sec:conclusion} concludes the paper.

\pagebreak

\section{Fairness Notions and Stakeholders}
\label{sec:stakeholders}

Different fairness notions are covered in the literature. While some may be better suited for LES studies, they might not be adequate for managing congestion or allocating HC at feeder-level. For this reason, we go over the different fairness notions, before covering how these shape fairness for stakeholders across different applications. 
The intrinsic fairness conception within all optimization frameworks is utilitarianism ($U$), which focuses on maximizing total welfare within a system \cite{chen2023}.
Note, $U$ can be interpreted as the efficient solution or efficiency maximizing solution. 
In fact, this is the efficiency baseline that must be balanced against various fairness notions, being positioned at one extreme of the fairness-efficiency trade-off spectrum \cite{sundar2025, hernandez2021}. In power grids, it leads to unfair treatment of specific nodes due to locational disparities within the feeder \cite{liu2020, rubbers2025, aydin2025a}.

\subsection{Fairness Notions}

In the aforementioned studies, the presented fairness concepts are derived from five big notions. The most widely used is the Egalitarian ($E$) notion, which focuses on ensuring that all participants receive an as equal as possible share of goods or burdens \cite{soares2024}. In DN, this is usually translated into every user receiving the same HC allocation or PV curtailment regardless of their location or total capacity \cite{zhan2024, rubbers2025}.

Despite this perspective being the typical fairness standard, resources may be allocated differently based on a pre-existing claim, right, entitlement, or specific physical attribute \cite{wei2015}. From the reviewed list of publications, Poudel et al. implements this notion on proportional burden sharing, namely curtailing PV systems as a portion of their total available PV output  \cite{poudel2023}. This criterion is covered by the Proportional ($P$) fairness perspective, where user allocations are typically proportional to their demand or their installed capacity  \cite{zhan2024, aydin2025a}. 

Similarly, focused on LESs, the Meritocratic ($M$) fairness notion generally uses the initial bill size to distribute the utility among participants \cite{couraud2025}. Here, the fundamental distinction made between $P$ and $M$ fairness notions lies in whether the focus is on equalizing the burden ($P$) or rewarding each member's contribution ($M$). Meritocratic fairness schemes ultimately entitle a larger share of the benefits to those who provide higher value to the system \cite{couraud2025}. Furthermore, they use the Shapley Value, a game theory concept which allocates profit based on each member's marginal contribution to a coalition, as the fair $M$ benchmark \cite{soares2024, gasca2025}. 

Bargaining ($B$) and Threshold ($T$) notions seek a middle point on the fairness-efficiency spectrum. 
On the one hand, $B$ notions, also denoted as trade-off fairness definitions, treat fairness as a deliberate balance between system efficiency and distributional equity \cite{rubbers2025}. Furthermore, a weighting parameter (such as $\mu$ in \cite{liang2025} or $K$ in \cite{rubbers2025} ) allows a negotiation about how much efficiency is sacrificed to reduce disparities. This fairness notion acknowledges the ingrained conflict where some efficiency must be sacrificed for higher degrees of fairness, as to ensure social acceptance or to prevent bill shocks \cite{schittekatte2020}.
On the other hand, $T$ notions are hybrid approaches that switch priorities between fairness and efficiency based on specified utility levels \cite{chen2023}. This threshold can be efficiency or fairness dependent. The first prioritizes fairness for participants whose utility falls below a certain limit, switching to $U$ once the PoF becomes too high or participants are no longer considered severly disadvantaged. Inversely, the latter begins with a utilitarian goal, switching to a fairness ensuring strategy if the resulting disparity goes beyond a given threshold \cite{chen2023}. For instance, in \cite{sun2022}, the VR mode selection, ruling between an optimal and a fair mode, is based on a predefined value for the acceptable economic benefits of the Distribution System Operator (DSO). Alternatively, bounded fairness criteria may also be implemented within this notion, setting upper and lower limits on HC distribution \cite{rubbers2025, lliuyacc-blas2025}.

\subsection{Stakeholder Representation}
Table \ref{tab:notionsstakeholders} presents the predominant fairness notions and stakeholders represented within the reviewed literature. 
\vspace{-8pt}
\begin{table}[H]
\centering
\small
\caption{Fairness Notions and Stakeholders}
\vspace{3pt}
\begin{tabular}{p{6.5cm}p{3.5cm}p{3cm}p{3cm}}
\hline
\textbf{Application} & \textbf{References} & \textbf{Notions} & \textbf{Stakeholders} \\
\hline
Congestion  &  \cite{liang2025} & $B$, $E$ & $A$, $C$, $D$ \\
Management   & \cite{liu2020, zhan2024} & $E$, $P$ & $C$, $D$ \\
(CM) & \cite{hernandez2021} & $E$, $P$ & $A$, $C$, $D$  \\
  &\cite{abdolahinia2025} & $E$, $T$ & $C$, $D$\\  \hline
General &  \cite{wei2015} & $E$, $P$ & $C$, $D$ \\
Optimization   & \cite{sundar2025} & $E$, $P$, $T$ & $C$, $D$ \\
(GO)   & \cite{chen2023} & $B$, $E$, $P$, $T$ & $A$, $C$, $D$, $R$\\  \hline
Hosting   & \cite{mahmoodi2021} & $E$, $P$ & $C$, $D$ \\
Capacity   & \cite{eldridge2023} & $E$, $P$ & $C$, $D$, $R$ \\
(HC)   & \cite{rubbers2025, lliuyacc-blas2025} & $B$, $E$, $T$ & $C$, $D$ \\
     & \cite{aydin2025a} & $E$, $P$, $T$ & $C$, $D$ \\ \hline
Local & \cite{casalicchio2022} & $E$, $M$ & $A$, $C$, $D$ \\
Energy & \cite{couraud2025} & $E$, $M$ & $C$, $D$ \\
System & \cite{abdolahinia2025} & $E$, $T$ & $C$, $D$ \\
(LES) & \cite{gasca2025} & $E$, $M$, $P$ & $A$, $C$ \\
 & \cite{soares2024} & $B$, $E$, $M$, $P$, $T$ & $A$, $C$, $D$ \\ \hline
Voltage  & \cite{hernandez2021} & $E$  & $A$, $C$, $D$\\
Regulation  & \cite{gerdroodbari2021} & $E$  & $C$, $D$ \\
 (VR) & \cite{hashmi2023} & $E$  & $C$, $D$, $R$ \\
  & \cite{poudel2023} & $P$  & $A$, $C$, $D$ \\
  & \cite{liu2020, gupta2025, gupta2025a} & $E$, $P$  & $C$, $D$  \\
  & \cite{gebbran2021} & $E$, $P$, $T$ & $A$, $C$, $D$, $R$ \\
  & \cite{sun2022} & $E$, $P$, $T$  & $C$, $D$ \\\hline
\multicolumn{4}{|p{16cm}|}{\footnotesize{Notions $\rightarrow$ E: egalitarian, P: proportional, B: bargaining, M: meritocratic, T: threshold. \quad Stakeholders $\rightarrow$ C: consumers, D: DSO, A: aggregator, R: regulator.}} \\\hline
\end{tabular}
\label{tab:notionsstakeholders}
\end{table}


Similarly to fairness notions, not all affected parties are equally represented across the literature. 
The overarching stakeholder across fairness-aware DN studies is the set of Consumers ($C$). In this study, $C$ encompasses both passive and active consumers, with and without DER such as solar PV or batteries (prosumers). They are modeled as independent agents seeking to minimize their electricity bills with the help of local energy management systems, or maximizing their welfare through LES schemes \cite{abdolahinia2025, gasca2025}. Ultimately, they are categorized by their physical location on a feeder, which dictates their susceptibility to voltage violations, and thus, curtailment \cite{hashmi2023}.
The DSO ($D$) is responsible for ensuring grid security, maintaining voltage and current within safe operational limits and managing congestion \cite{lliuyacc-blas2025}. It acts as a market clearer who provides the technical grid data and design price signals to influence consumer behavior \cite{abdolahinia2025}. In fairness studies, it is this organization's responsibility to ensure equitable grid access and fair curtailment is applied \cite{hernandez2021}.
The Aggregator, hereby $A$, plays the role of bundling small DER assets to enable their participation in electricity markets, acting as a bridge between $C$ and $D$, controlling their portfolio disaggregatedly via specific reward mechanisms \cite{poudel2023, tsaousoglou2021}. 
Finally, Regulators ($R$) establish the legal and economic frameworks defining how markets and energy communities operate \cite{casalicchio2022, gasca2025}. By balancing economic efficiency with social fairness, they design network tariffs that prevent aforementioned bill shocks for $C$ (consequently $A$), all the while ensuring cost recovery possibilities for $D$ \cite{schittekatte2020}. Effectively, and referencing the previous distinction within $C$, regulatory bodies protect passive consumers from cost shifts caused by active prosumers by identifying proxies for guaranteeing fair burden-sharing \cite{schittekatte2020}.

\pagebreak
\section{Fairness and Price of fairness}
\label{sec:pof}
\vspace{-4pt}
A consumer's position on a grid's topology primarily dictates their hierarchy in business as usual (BAU) schemes, where efficiency-centric schemes prioritize those who are closer to the substations.
While the physics of the grid makes locational disparity a natural outcome of efficient operation, regulatory mandates require DSOs to provide a "level-playing field" for all consumers, necessitating fairness-aware controls \cite{hashmi2023}. For instance, metrics such as the LCG or the Net Benefit Index (NBI) measure the financial sacrifice prosumers make due to their location on the grid due to BAU control methods \cite{liu2020, hashmi2023}. Alternatively, physical (e.g. PHI and EEI indexes from \cite{liu2020}) are also used to measure the extent of locational disparity. However, implementing schemes that incorporate fairness comes at a cost, that cost is represented through the PoF.

The formulation of the PoF, as in \cite{bertsimas2011}, provides a simple comparison between the utilitarian ($U$) outcome, with the results of any fairness-aware scheme ($F$), $\text{PoF} = \frac{U - F}{U}$. Thus, the PoF encompasses the efficiency that is sacrificed towards achieving a fair resource allocation, be it residual HC, PV curtailment or total power losses \cite{abdolahinia2025, rubbers2025, lliuyacc-blas2025, sun2022, gupta2025a}. Furthermore, some metrics are intrinsically tied to the PoF. For instance, the aforementioned $MI$ and $F$ indices allow for a grasp of how much efficiency would have to be satisfied to attain the specific ideal contribution, as they are ultimately a difference between an efficiency-oriented scheme and the most fair scenario for that meritocratic criterion \cite{soares2024, couraud2025}.

Stakeholder-wise, while regulators settle on an acceptable PoF threshold, DSO's are typically the ones incurring the PoF, manifesting as higher operational costs or lower total renewable integration \cite{sun2022, schittekatte2020, tsaousoglou2021}. However, consumers, namely the ones at more resilient nodes, also pay the PoF. This is sometimes carried out through a reduction of their individual output to ensure those at the end of the feeder have equal harvesting opportunities \cite{mahmoodi2021}.  

\pagebreak

\section{Fairness Integration}
\label{sec:metrics}

Firstly, we define the terms which will be used in the metrics' formulations. Therefore, $u_i$ is defined as the allocated utility to each member $i$, with $N$ representing the total number of members/individuals ($i \in N$). Within the $U$ fairness notion, its sum, $\sum^N_{i=1}u_i$, is the core fairness metric, with a desired fair value based on whether the objective is to maximize the total social welfare, or minimize the shared burden. $\overline{u}$ is the median or average of utilities. For the McLoone Index specifically, $I(u)$ is the set of indices of utilities at or below the median \cite{chen2023}. Overall, utility can be assigned within three overarching scopes: benefit distribution, (dis)comfort, and pricing \cite{soares2024}.

\subsection{Metrics for DN fairness}
\label{sec:metricssub}
From inequality measures to methods combining fairness and efficiency, with and without thresholds, there exists a multitude of metrics for measuring fairness, based on the desired fairness notion. From \cite{chen2023}, one obtains a comprehensive overview of the most common ones. However, as the guide does not focus on DN fairness, not all of them are interesting when formulating fairness in this specific context. Contrary to group parity measures, which do not target our intended scope, certain inequality measures are employed within DN fairness studies. 
\begin{table}[H]
\centering
\small
\caption{Metrics for DN Fairness}
\vspace{3pt}
\begin{tabularx}{\textwidth}{m{6cm}Xm{2.8cm}m{2.8cm}}
\hline
\textbf{Metric} & \textbf{Formula} & \textbf{Range} & \textbf{Fair Value} \\
\hline
Epsilon-fairness ($\varepsilon$) &
$(1 - \varepsilon + \varepsilon\sqrt{N})\|\mathbf{p}\|_2 \le \|\mathbf{p}\|_1$ &
$[0, 1]$ &
$1$ \\
\hline
Jain's Index ($J$) &
$\frac{(\sum^N_i u_i)^2}{N \sum_i^N u_i^2}$ &
$[0,1]$ &
$1$ \\ 
\hline
Min-Max Ratio ($MiM$) &
$\frac{\min_iu_i}{\max_iu_i}$ &
$[0,1]$ &
$1$ \\ 
\hline
McLoone Index ($M_L$) &
$\frac{1}{|I(u)|\,\overline{u}} \sum^N_{i\in I(u)} u_i$ &
$[0,1]$ &
$1$ \\
\hline
Gini coefficient ($G$) &
$\frac{1}{2\overline{u}N^2}\sum_{i,j}^N|u_i - u_j|$ &
$[0,1]$ &
$0$ \\ 
\hline
Hoover Index ($H$) &
$\frac{1}{2N\overline{u}} \sum^N_i |u_i-\overline{u}|$ &
$[0,1]$ &$0$ \\ 
\hline
F Fairness Index ($F$) &
$\sum^N_{i=1} \left| \frac{B_i}{\sum^N_{i=1}B_i}-\frac{S_i^*}{\sum^N_{i=1}S_i^*}\right|$ &
$[0,+\infty)$ &
$0$ \\ 
\hline    
Meritocratic Index ($MI$) &
$\sqrt{\frac{1}{N}\sum_i^N (u_i - \hat{u}_i)^2}$ &
$[0,+\infty)$ &
$0$ \\ 
\hline
Alpha-fairness ($\alpha$) &
$\begin{cases}
\frac{1}{1-\alpha}\sum^N_iu_i^{\alpha-1} \quad \text{for }  \alpha \neq 1\\[6pt]
\sum_i^N\log(u_i )\qquad \text{for } \alpha =1  
\end{cases}$ &
$[0, +\infty)$ &
$*$\\
\hline
Social Welfare ($u$) &
$ \sum_{i=1}^N u_i$ &
$[0,+\infty)$ &
$**$ \\  
\hline
\multicolumn{4}{p{\textwidth}}{\footnotesize
* See Section \ref{sec:metricssub} for fair value interpretation. 
** Explained above, in \ref{sec:metrics}.
}\\
\hline
\end{tabularx}
\label{tab:metrics}
\end{table}

For instance, Jain's (Fairness) Index ($J$) is a widely adopted $E$ metric, originally developed for telecommunication networks, which proves itself useful in DNs recurrently \cite{chen2023}. For this reason, Table \ref{tab:metrics}, compiles commonly used metrics, as well as some that may prove useful in incorporating fairness within power distribution systems.

For the $\varepsilon$-fairness metric, as employed in \cite{aydin2025a}, $\mathbf{p}$ represents the vector of utilised resources, e.g. the HC, and N the number of agents, in this case, the nodes with generation. This metric implements a $B$ fairness notion, leveraging the L1-L2 norm inequality, as the chosen value for $\varepsilon$ constrains the fairness-efficiency trade-off to at least that specific degree of fairness. This metric improves on the more traditionally $B$ metric $\alpha$-fairness, which, unlike $\varepsilon$-fairness, ranges from 0 to infinity \cite{sundar2025}. In fact, $\alpha$-fairness covers multiple fairness notions: $\alpha=0$ represents the $U$ criterion, $\alpha=1/N$ covers $P$ and $B$ fairness (known as the Nash bargaining solution), covering a purely maximin criterion as the parameter grows towards infinity \cite{chen2023}. This last criterion focuses on improving the allocation of the worse-off, as it is based on the Rawlsian fairness principle, a less strict form of $E$ fairness \cite{chen2023}. Despite this, both $\varepsilon$ and $\alpha$ are vulnerable to inadequate use, as it is not clear what value to use, and how much a change in this value alters the fair allocation of resources.

From LES literature, two metrics, namely the F Fairness index ($F$) and the Meritocratic Index ($MI$) with similar fairness approaches are considered. In $F$'s formulation, $B_i$ is the bill or income of prosumer $i$ when using other trading mechanisms, and $S^*_i$ is the bill or income of prosumer $i$ when using the SV \cite{soares2024, long2019}. Likewise, for the $MI$, $\hat{u}_i$ is member $i$'s ideal meritocratic share, \cite{couraud2025}. Contrary to the $F$ index, the ideal notion is not the SV, but instead, the product of the member’s normalized contribution $\frac{C_i}{\sum_j C_j}$ and the total community savings $\sum_j u_j$ \cite{couraud2025}. The reason for this difference is presented in Section \ref{sec:mathmodeling}. Intuitively, the fair value for the social welfare metric will vary based on wether the objective is to maximise the allocated utility, for which the fair case is the highest attainable case, or minimize the shared burden, where the opposite takes place.


\subsection{Mathematical Modeling}
\label{sec:mathmodeling}
Two of the metrics in Table \ref{tab:metrics} are intrinsically tunable to the desired degree of fairness: $\alpha$ and $\varepsilon$. Despite not being convex as $\varepsilon$ is, and being in fact irreducibly non-linear, $\alpha$ is concave for all $\alpha\geq 0$ \cite{sundar2025}. Hence, any local optimum is a global optimum if the feasible set is convex \cite{chen2023}. Except for these two metrics, a minimum level of fairness is not necessarily forced onto the resource allocation. However, it is possible to set a minimum acceptable fairness level as a separate optimisation constraint, with a defined minimum $J$ value, for example \cite{abdolahinia2025}. In fact, as the $\varepsilon$-fairness is a direct proxy for the $J$ index (bounded from below), its use incorporates this extra constraint into the fairness metric itself \cite{aydin2025a, sundar2025}. The $J$ index, non-linear and not inherently convex, is a strictly monotone function of the coefficient of variation, which can be reformulated into a convex objective function \cite{chen2023}. Alternatively, $J$ may be linearised into a constraint for market-clearing problems with the use of a Taylor series approximation \cite{abdolahinia2025}.

Regarding $G$, $H$, $MiM$ and $M_L$, despite being mathematically non-linear, they are readily linearised for linear programming (LP) or mixed-integer linear programming (MILP). For $G$ and $MiM$, as it is a ratio of affine functions, it can be linearised using change of variable techniques \cite{chen2023}. $H$ can be minimized using a linear model, as it is proportional to the relative mean deviation \cite{chen2023}. $M_L$ requires 0-1 integer variables, in order to represent the median, and a fractional objective function which can be linearised using an MILP model \cite{chen2023}.

Lastly, the computational advantage of $MI$ over the $F$, is that while SV gets severely computationally demanding for larger volumes, $MI$ is calculated as an RMS Error (RMSE), which when minimized leads to convex quadratic programming \cite{couraud2025, gerdroodbari2021}.

\pagebreak

\section{Conclusions}
\label{sec:conclusion}

This study reviewed how the different fairness notions represent stakeholders in DNs. We covered how locational disparity impacts consumers, and how the efficiency sacrifice towards a fair resource allocation is taken into account in recent studies. The metrics used to integrate fairness in DNs were also presented, comparing their mathematical formulations and computational tractability. Overall, this review proposes a structured overview of the different approaches for formulating fairness with explicit efficiency tradeoffs, incorporating the price of fairness (PoF) notion, enabling transparent, equitable and efficient strategies for future DN planning and operation.

Future work on fairness integration should focus on guaranteeing that the PoF is clear and that locational disparities are actively tackled to achieve transparent and fair decision-making.

\section*{Acknowledgments}
This work was partially funded by the KU Leuven-funded C2-project FlexIQ (C2M/24/028).

\pagebreak

\bibliographystyle{IEEEtran}
\bibliography{references.bib}

\end{document}